\documentclass{article}
\usepackage[preprint]{spconf}
\usepackage{amsmath,graphicx}
\usepackage[hidelinks]{hyperref}
\usepackage{booktabs} %
\usepackage{subcaption}
\usepackage{xcolor, colortbl}
\usepackage{xspace}
\usepackage{multirow}
\usepackage{cite}

\usepackage{amsfonts}

\title{Reproducing Whisper-Style Training Using an Open-Source Toolkit and Publicly Available Data}
\name{
    \begin{tabular}[c]{@{}c@{}c@{}c@{}c@{}c@{}}
        Yifan Peng$^1$, Jinchuan Tian$^1$, Brian Yan$^1$, Dan Berrebbi$^1$, Xuankai Chang$^1$, Xinjian Li$^1$, Jiatong Shi$^1$, \\
        Siddhant Arora$^1$, William Chen$^1$, Roshan Sharma$^1$, Wangyou Zhang$^{1,2}$, Yui Sudo$^3$, Muhammad Shakeel$^3$, \\
        Jee-weon Jung$^1$, Soumi Maiti$^1$, Shinji Watanabe$^1$
    \end{tabular}
}
\address{$^1$Carnegie Mellon University, USA\\
$^2$Shanghai Jiao Tong University, China\\
$^3$Honda Research Institute Japan, Japan
}

\copyrightnotice{979-8-3503-0689-7/23/\$31.00~\copyright2023 IEEE}
\begin{document}
\ninept
\maketitle
\begin{abstract}
Pre-training speech models on large volumes of data has achieved remarkable success. OpenAI Whisper is a multilingual multitask model trained on 680k hours of supervised speech data. It generalizes well to various speech recognition and translation benchmarks even in a zero-shot setup. However, the full pipeline for developing such models (from data collection to training) is not publicly accessible, which makes it difficult for researchers to further improve its performance and address training-related issues such as efficiency, robustness, fairness, and bias. This work presents an Open Whisper-style Speech Model (OWSM), which reproduces Whisper-style training using an open-source toolkit and publicly available data. OWSM even supports more translation directions and can be more efficient to train. We will publicly release all scripts used for data preparation, training, inference, and scoring as well as pre-trained models and training logs to promote open science.\footnote{\url{https://github.com/espnet/espnet}}
\end{abstract}

\section{Introduction}

Large-scale Transformers~\cite{transformer} have garnered significant attention in natural language processing (NLP)~\cite{gpt3,rae2021scaling,chowdhery2022palm,zhang2022opt,touvron2023llama,OpenAI2023GPT4TR}. These models, trained on extensive datasets, have showcased remarkable emergent capabilities in diverse downstream tasks. Notably, the application of similar pre-training techniques has also found success in the domain of speech processing. Self-supervised learning (SSL) techniques have demonstrated impressive achievements~\cite{wav2vec2, hubert, xlsr, superb, ssl-review, ssl-for-asr, ssl-for-slu}. Furthermore, large-scale supervised learning has emerged as a promising avenue for the development of universal speech models capable of performing multiple speech tasks within a single model~\cite{openai-whisper, speechstew, 9687871, google-usm}. OpenAI Whisper~\cite{openai-whisper} is a series of multilingual multitask models trained on 680k hours of labeled speech data which is carefully curated from diverse sources on the Internet. 

Despite the release of pre-trained Whisper models and inference code, the comprehensive pipeline for model development (from data preparation to training) remains inaccessible to the public, which has been a common situation for large language models (LLMs). This limitation engenders several concerns. 
Firstly, the utilization of pre-trained models on novel benchmarks has the potential risk of data leakage, as users are deprived of knowledge regarding the actual training data.
Secondly, researchers face significant difficulties in comprehending the underlying mechanisms and elucidating methods for enhancing the model's performance, given their lack of access to the training dynamics.
Thirdly, the absence of access to the complete model development pipeline poses notable challenges in effectively tackling issues related to robustness, fairness, bias, and toxicity, all of which frequently arise as a result of the data and training procedure~\cite{liang2021towards, wang2023robustness, bubeck2023sparks}.

Recently, there has been a concerted effort to foster open science in the realm of LLM research by advocating for the release of complete training pipelines~\cite{zhang2022opt}. Inspired by this, we present the Open Whisper-style Speech Model (OWSM)\footnote{OWSM is pronounced as ``awesome''.}, which reproduces Whisper-style training using an open-source toolkit and publicly available data. OWSM follows the design of Whisper~\cite{openai-whisper} to support essential tasks such as language identification (LID), multilingual automatic speech recognition (ASR), and utterance-level segmentation. Notably, OWSM also exhibits several technical novelties. It is designed to support any-to-any speech translation as opposed to solely any-to-English translation (see Section~\ref{subsec:exp-st} for results). OWSM also adopts multiple strategies to enhance the efficiency (see Section~\ref{subsec:trainingtips} for discussions).

We will provide reproducible recipes encompassing the entire pipeline, including data preparation, training, inference, and scoring. Furthermore, we will release pre-trained models and training logs, enabling researchers to delve into the specifics of the training process and gain valuable insights for their own investigations. 
While OWSM shows competitive or even superior performance compared to Whisper in certain benchmarks, it is essential to clarify that our objective is not to engage in a comprehensive competition with Whisper. The scope of our endeavor is constrained by the fact that our largest dataset comprises only a quarter of the training set used by Whisper, and our resource limitations restrict us from conducting multiple trial runs. Instead, by sharing these resources, we aim to promote transparency and facilitate progress and advancements in the field of large-scale pre-training for speech processing.

\section{Whisper-style training}

\begin{figure*}[t]
\centering
\includegraphics[width=0.95\linewidth]{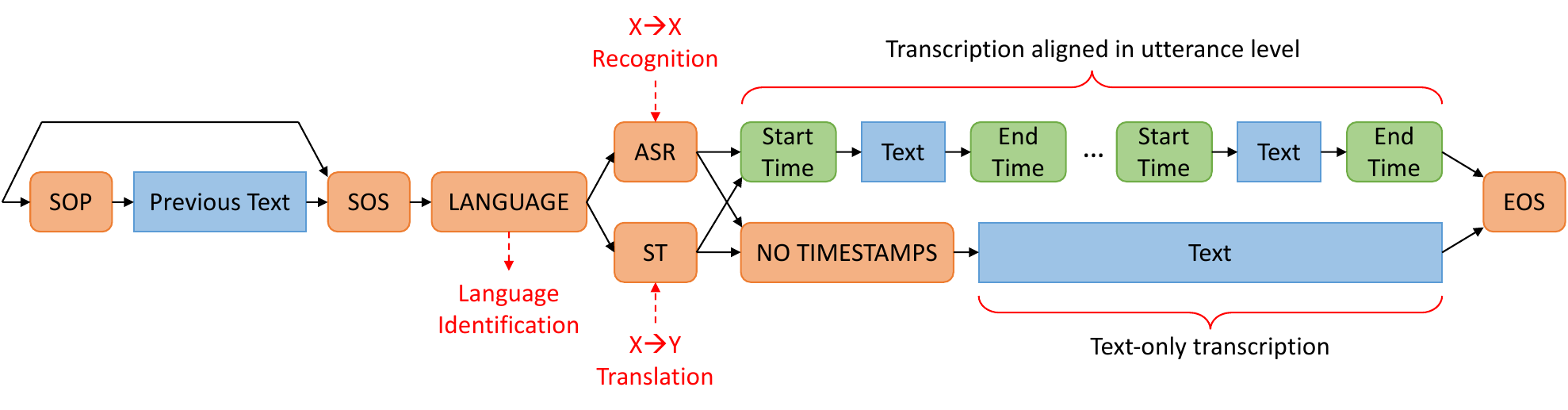}
\caption{Multitask data format used by our OWSM, which mostly follows OpenAI Whisper~\cite{openai-whisper}. Different speech processing tasks are represented in a unified format, which can be predicted by an autoregressive decoder. Note that OWSM is designed to support any-to-any speech-to-text translation, whereas Whisper can only perform any-to-English translation. Blue boxes denote standard text tokens, while orange and green boxes are special tokens. SOP, SOS, and EOS represent start-of-prompt, start-of-sentence, and end-of-sentence, respectively.}
\label{fig:overview}
\end{figure*}

\begingroup
\setlength{\tabcolsep}{3pt}
\begin{table}[t]
  \caption{Details of data, model architectures, and training configurations. We gradually increase data and model sizes from v1 to v3. The model configurations of OWSM v1 and v3 match those of Whisper small and medium, respectively. Although OWSM v3 covers more languages than Whisper, our data size remains significantly smaller, making our task much more challenging. $^*$Our v3 model is initialized with the pre-trained v2 to reduce training time (see Section~\ref{subsec:trainingtips}).}
  \label{tab:data-model-stats}
  \centering
  \resizebox {\linewidth} {!} {
  \begin{tabular}{lcccccc}
    \toprule
     & \multicolumn{3}{c}{OpenAI Whisper} & \multicolumn{3}{c}{OWSM (ours)}\\
     \cmidrule(lr){2-4}
     \cmidrule(lr){5-7}
     & small & medium & large & v1 & v2 & v3$^*$\\
    \midrule
    \multicolumn{7}{l}{\textit{Data}}\\
    \quad Total hours (k) & \multicolumn{3}{c}{680} & 38 & 129 & 180 \\
    \qquad - English ASR & \multicolumn{3}{c}{438} & 22 & 67 & 73 \\
    \qquad - Multilingual ASR & \multicolumn{3}{c}{117} & 1 & 22 & 67 \\
    \qquad - Translation & \multicolumn{3}{c}{125} & 15 & 40 & 40 \\
    \quad Languages & \multicolumn{3}{c}{99} & 22 & 23 & 151 \\
    \quad BPE vocabulary size & \multicolumn{3}{c}{51,865} & 20k & 50k & 50k \\
    \midrule
    \multicolumn{7}{l}{\textit{Model architectures}}\\
    \quad Parameters (M) & 244 & 769 & 1550 & 272 & 712 & 889\\
    \quad Hidden size & 768 & 1024 & 1280 & 768 & 1024 & 1024 \\
    \quad Layers & 12 & 24 & 32 & 12 & 18 & 24 \\
    \quad Attention heads & 12 & 16 & 20 & 12 & 16 & 16 \\
    \quad Time resolution (ms) & 20 & 20 & 20 & 20 & 40 & 40 \\
    \midrule
    \multicolumn{7}{l}{\textit{Training configurations}}\\
    \quad Batch size & \multicolumn{3}{c}{256} & \multicolumn{3}{c}{256} \\
    \quad Total updates & \multicolumn{3}{c}{1,048,576} & 300k & 500k & 470k \\
    \quad Warmup updates & \multicolumn{3}{c}{2048} & 10k & 20k & 10k \\
    \quad Learning rate & 5e-4 & 2.5e-4 & 1.75e-4 &  1e-3 & 5e-4 & 2.5e-4 \\
    \quad Optimizer & \multicolumn{3}{c}{AdamW} & \multicolumn{3}{c}{AdamW} \\
    \quad Joint CTC weight & \multicolumn{3}{c}{NA} & \multicolumn{3}{c}{0.3}  \\
    \bottomrule
  \end{tabular}
  }
\end{table}
\endgroup

\subsection{Multitask data format}
\label{subsec:multitask-data-format}

OpenAI Whisper~\cite{openai-whisper} employs a single sequence-to-sequence model to perform multiple speech processing tasks, including LID, multilingual ASR, any-to-English ST, and utterance-level segmentation. Our OWSM mostly follows this design, but extends it to potentially support any-to-any ST. Figure~\ref{fig:overview} illustrates the multitask data format. Data samples from different tasks are represented in a unified format, which can be predicted by the decoder in an autoregressive manner. Specifically, each sample is converted to a sequence of tokens with two segments separated by special tokens. The first segment (before ``SOS'') is an optional text prompt used as a condition, while the second segment is the actual target. The target starts with a special token denoting the language of the input speech. Then, it uses a task token to distinguish between ASR and ST. There is a separate ST token for each target language, which enables translation to any language. Finally, it appends the text transcription either with or without utterance-level timestamps. All timestamps are quantized and represented as special tokens.

\subsection{Data preparation}
\label{subsec:dataprep}
Whisper is pre-trained on 680k hours of labeled audio data sourced from the Internet, which is not publicly accessible. To construct a speech dataset for large-scale supervised learning, we combine training sets from various publicly available ASR and ST corpora. These diverse corpora encompass a wide range of speaking styles, recording environments, and languages.
Our datasets are prepared using an open-source toolkit, ESPnet~\cite{espnet}. However, OWSM is trained on long-form audio data, which deviates from previous recipes in ESPnet. Consequently, we have developed new data preparation scripts tailored specifically for Whisper-style training. We concatenate consecutive utterances within the same long talk based on their original timestamps. Each long-form utterance is limited to a maximum duration of 30 seconds. During training, all utterances are padded to precisely 30 seconds, optimizing the utilization of computational resources.

To date, we have developed three versions at different scales, denoted as OWSM v1, v2, and v3 in Table~\ref{tab:data-model-stats}. Our largest dataset, v3, comprises 180k hours of labeled audio data. This constitutes approximately one quarter of the total data employed by OpenAI Whisper in its training process~\cite{openai-whisper}.
The individual datasets utilized by our models are listed below:
\begin{itemize}
    \item OWSM v1: AISHELL-1~\cite{aishell-corpus}, CoVoST2~\cite{covost2}, GigaSpeech~\cite{gigaspeech}, LibriSpeech~\cite{librispeech-corpus}, MuST-C~\cite{must-c}, SPGISpeech~\cite{spgispeech}, and TEDLIUM3~\cite{tedlium3}.
    \item OWSM v2: all data in v1, GigaST~\cite{gigast}, Multilingual LibriSpeech~\cite{pratap2020mls}, and WenetSpeech~\cite{wenetspeech}.
    \item OWSM v3: all data in v2, AIDATATANG~\cite{aidatatang}, AMI~\cite{ami-corpus}, Babel~\cite{babel}, Common Voice~\cite{commonvoice}, Fisher (Switchboard)~\cite{swbd-corpus}, Fisher Callhome Spanish~\cite{fisher-callhome}, FLEURS~\cite{FLEURS}, Googlei18n\footnote{Resources 32, 35, 36, 37, 41, 42, 43, 44, 52, 53, 54, 61, 63, 64, 65, 66, 69, 70, 71, 72, 73, 74, 75, 76, 77, 78, 79, and 86 from \url{openslr.org}.}, KsponSpeech~\cite{ksponspeech}, MagicData~\cite{magicdata}, ReazonSpeech~\cite{reazonspeech}, Russian Open STT~\cite{ru-open-stt}, VCTK~\cite{vctk}, VoxForge~\cite{voxforge}, VoxPopuli~\cite{voxpopuli}, and WSJ~\cite{wsj}.
\end{itemize}

\subsection{Model architectures}

OWSM follows Whisper to utilize a Transformer encoder-decoder architecture~\cite{transformer}, where the encoder and decoder have the same number of layers. However, OWSM additionally employs a joint CTC loss for ASR targets~\cite{joint-ctc-attention}, which was empirically shown to stabilize our training process. The input waveforms are converted to 80-dimensional log Mel filterbanks with a window length of 25ms and a hop length of 10ms. The extracted features are augmented using SpecAugment~\cite{specaugment} and normalized by their global mean and variance. The features are then processed by a two-dimensional convolution module to reduce the sequence length. OpenAI Whisper~\cite{openai-whisper} always downsamples the sequence by 2, resulting in a time resolution of 20ms. Our OWSM v2 and v3 perform 4 times downsampling, which further improves efficiency. The detailed configurations of Transformer encoder and decoder layers are summarized in Table~\ref{tab:data-model-stats}. OWSM v1 and v3 use the same configurations as Whisper small and medium, respectively, while OWSM v2 is slightly smaller than v3.~\footnote{OWSM has slightly more parameters than Whisper under the same configuration, because the ESPnet model has a larger convolution downsampling module and does not share the input embedding and output projection in its decoder.}

For inference, OpenAI Whisper implements both greedy decoding and beam search with temperature fallback. The latter is a complicated procedure relying on many heuristics and hyperparameters such as beam sizes, temperatures, log probability threshold, and gzip compression rate threshold. Our OWSM utilizes the ESPnet framework~\cite{espnet}, thereby ensuring compatibility with various decoding algorithms originally supported by ESPnet, including greedy search, beam search, and joint CTC/attention decoding (for ASR only)~\cite{joint-ctc-att-decoding}.

\subsection{Training details}

OWSM is implemented in ESPnet~\cite{espnet} based on PyTorch~\cite{pytorch}. Table~\ref{tab:data-model-stats} compares the training hyperparameters of different models. OWSM uses the same batch size as Whisper, but the number of total updates is smaller. OWSM is trained on NVIDIA A100 (40GB) GPUs. Each GPU takes two samples, and gradient accumulation is applied whenever necessary to ensure the total batch size is 256. Specifically, OWSM v1 is trained for around 7 days on 32 A100 GPUs and OWSM v2 and v3 are trained for around 10 days on 64 A100 GPUs. After training, five checkpoints with the highest validation accuracies are averaged to generate the final checkpoint.

\begingroup
\begin{table}[t]
  \caption{WER \% ($\downarrow$) of English ASR using greedy search. OpenAI Whisper uses 438k hours of English ASR data, while OWSM uses at most 73k hours. As shown in Table~\ref{tab:data-model-stats}, the configurations of OWSM v1 and v3 match those of Whisper small and medium, respectively. Whisper large is significantly larger, so it is not included in the comparison. $^\dagger$The larger degradation of OWSM v3 on WSJ is likely caused by inconsistent case and punctuation of the training data (see the last paragraph in Section~\ref{subsec:trainingtips}). The gray color means OWSM is better than Whisper small and medium.}
  \label{tab:en-asr}
  \centering
  \resizebox {\linewidth} {!} {
  \begin{tabular}{cccccc}
    \toprule
    \multirow{2}{*}{Dataset} & \multicolumn{2}{c}{OpenAI Whisper} & \multicolumn{3}{c}{OWSM (ours)}\\
    \cmidrule(lr){2-3}
    \cmidrule(lr){4-6}
    & small & medium & v1 & v2 & v3\\
    \midrule
    Common Voice en & 15.7 & \textbf{11.9} & 20.1 & 14.4 & 14.5 \\
    FLEURS en & 9.6 & \textbf{6.4} & 13.2 & 10.9 & 10.9 \\
    LibriSpeech test-clean & 3.3 & 2.8 & 5.4 & \cellcolor{gray!50} \textbf{2.2} & \cellcolor{gray!50} 2.7 \\
    LibriSpeech test-other & 7.7 & 6.5 & 10.9 & \cellcolor{gray!50} \textbf{5.1} & \cellcolor{gray!50} 6.0 \\
    Switchboard eval2000 & 22.2 & 19.4 & 28.7 & 20.4 & \cellcolor{gray!50} \textbf{17.2} \\
    TEDLIUM test & \textbf{4.6} & 5.1 & 6.6 & \cellcolor{gray!50} \textbf{4.6} & 4.8 \\
    VoxPopuli en & 8.5 & \textbf{7.6} & 14.2 & 10.3 & 9.2 \\
    WSJ eval92 & 4.3 & \textbf{2.9} & 4.3 & 3.7 & 13.4$^\dagger$ \\
    \bottomrule
  \end{tabular}
  }
\end{table}
\endgroup

\begingroup
\setlength{\tabcolsep}{5pt}
\begin{table*}[t]
  \caption{WER/CER \% ($\downarrow$) of multilingual ASR using greedy search. Training data sizes (in hours) are also provided. The gray color means OWSM is better than Whisper small and medium.}
  \label{tab:multilingual-asr}
  \centering
  \resizebox {0.9\linewidth} {!} {
  \begin{tabular}{cccccccccccc}
    \toprule
    \multirow{2}{*}{Dataset} & \multirow{2}{*}{Language} & \multirow{2}{*}{Metric} & \multicolumn{3}{c}{OpenAI Whisper} & \multicolumn{2}{c}{OWSM v1} & \multicolumn{2}{c}{OWSM v2} & \multicolumn{2}{c}{OWSM v3}\\
    \cmidrule(lr){4-6}
    \cmidrule(lr){7-8}
    \cmidrule(lr){9-10}
    \cmidrule(lr){11-12}
    & & & hours & small & medium & hours & result & hours & result & hours & result\\
    \midrule
    \multirow{8}{*}{Multilingual LibriSpeech} & English & \multirow{8}{*}{WER} & 438k & 9.1 & 10.2 & 22k & 13.7 & 67k & \cellcolor{gray!50} \textbf{6.7} & 73k & \cellcolor{gray!50} 7.4 \\
    & Spanish & & 11k & 9.1 & \textbf{6.1} & 0.1k & 37.2 & 1.0k & 11.7 & 2.0k & 11.7 \\
    & French & & 10k & 13.6 & \textbf{9.7} & 0.3k & 41.8 & 1.3k & 13.0 & 2.5k & 14.1 \\
    & German &  & 13k & 11.5 & \textbf{8.1} & 0.2k & 43.3 & 2.2k & 11.8 & 3.7k & 11.9 \\
    & Dutch &  & 2.1k & 18.2 & \textbf{12.2} & 0.007k & 78.7 & 1.6k & 16.9 & 1.7k & 17.7 \\
    & Italian & & 2.6k & 21.3 & \textbf{15.6} & 0.04k & 54.9 & 0.3k & 23.1 & 0.7k & 24.5 \\
    & Portuguese & & 8.6k & 13.8 & \textbf{8.9} & 0.009k & 90.9 & 0.2k & 31.8 & 0.3k & 28.2 \\
    & Polish &  & 4.3k & 12.5 & \textbf{6.8} & 0 & NA & 0.1k & 89.7 & 0.3k & 37.0 \\
    \midrule
    AISHELL-1 & Chinese & \multirow{4}{*}{CER} & 23k & 25.1 & 15.7 & 0.2k & 22.6 & 15k & \cellcolor{gray!50} \textbf{5.9} & 16k & \cellcolor{gray!50} 7.1 \\
    KsponSpeech eval-clean & \multirow{2}{*}{Korean} & & \multirow{2}{*}{8k} & 24.0 & \textbf{17.6} & \multirow{2}{*}{0} & \multirow{2}{*}{NA} & \multirow{2}{*}{0} & \multirow{2}{*}{NA} & \multirow{2}{*}{1.0k} & 20.5 \\
    KsponSpeech eval-other & & & & 15.4 & \textbf{12.8} & & & & & & 22.6 \\
    ReazonSpeech & Japanese & & 7k & 32.5 & 25.3 & $\approx$0 & NA & $\approx$0 & NA & 19k & \cellcolor{gray!50} \textbf{11.3} \\
    \bottomrule
  \end{tabular}
  }
\end{table*}
\endgroup

\subsection{Challenges and training tips}
\label{subsec:trainingtips}

Large-scale distributed training presents significant challenges, particularly when the computation budget is limited. As we scale up from a few thousand hours of data to nearly 200 thousand hours, we have encountered a range of issues. Here, we discuss some of the challenges and provide valuable training tips to help overcome these obstacles effectively. We will release our scripts that support these techniques.

\textbf{Time resolution:} Whisper employs a time resolution of 20ms within its encoder module, resulting in a sequence length of 1500 for 30-second inputs. This significantly increases GPU memory consumption and makes training slower and more difficult. In contrast, contemporary state-of-the-art ASR and ST models~\cite{conformer, branchformer, ebranchformer, squeezeformer} adopt larger downsampling rates. Starting from OWSM v2, we have adopted a time resolution of 40ms, effectively reducing the sequence length and mitigating the associated computational demands. We have also found that a shorter sequence length facilitates easier convergence of the model.

\textbf{Joint ASR CTC loss:} In our preliminary experiments, we observed suboptimal convergence of the attention-based encoder-decoder model trained on multiple tasks and diverse data. Incorporating a joint ASR CTC loss~\cite{joint-ctc-attention} to the encoder output can stabilize training and expedite convergence.

\textbf{Warm initialization:} When training our largest model, OWSM v3, we employ a warm initialization technique by leveraging the pre-trained OWSM v2. Specifically, the first 18 layers of OWSM v3 are initialized with v2 (which has precisely 18 layers), whereas the remaining 6 layers are initialized randomly. This v3 model converges much faster than training from scratch. However, it remains to be investigated whether a warm initialization adversely affects the final performance of the model.

\textbf{Memory and efficiency issues:} We have developed several strategies to address memory and efficiency issues caused by large data. 
To train the BPE tokenization models using SentencePiece~\cite{sentencepiece}, we randomly select 10 million text transcriptions instead of using the whole set to reduce memory usage.
For training, the entire text file is too large to be distributed across different workers. We partition the training set into 5 to 12 non-overlapping subsets and use multiple data iterators to construct mini-batches. We further filter out samples with extremely long transcriptions (e.g., greater than 600 tokens including both prompt and target) which are caused by incorrect alignments in the original corpus (e.g., Common Voice). Without such filtering, the training will occasionally encounter out-of-memory errors.
Additionally, we validate intermediate checkpoints using only 10\% of the full validation set. This might generate slightly inaccurate estimates of the actual performance, but it significantly reduces the validation time and thus allows for more frequent validation and checkpoint saving, which is crucial for large-scale distributed training. 
In fact, we encountered various failures mainly due to file system or communication errors, and we had to manually resume from previous checkpoints.

\textbf{Inconsistent case and punctuation.} Our training data is gathered from many public corpora. Some of them provide raw transcripts in true case with punctuation, but the others only provide normalized transcripts in lower or upper case without any punctuation. During inference, we find that OWSM models are so powerful that they are able to recognize the corpus and generate outputs that are consistent with the training data format. For example, the training data of WSJ is in upper case. When tested on WSJ test sets, OWSM also mostly generates text in upper case. Since only a very small portion of training data is in upper case, OWSM v3 performs poorly on WSJ (see \autoref{tab:en-asr}). In the future, we will normalize the text to address this issue. Note that this analysis demonstrates the benefit of using public data and open-source code, without which we cannot discover such issues.

\begingroup
\begin{table}[t]
  \caption{WER \% ($\downarrow$) of long-form ASR on the TEDLIUM2 test set. Unsegmented long talks are transcribed in chunks of 30 seconds. It is shifted based on predicted timestamps.}
  \label{tab:longform-asr}
  \centering
  \resizebox {0.82\linewidth} {!} {
  \begin{tabular}{ccccc}
    \toprule
    \multirow{2}{*}{Beam size} & \multicolumn{2}{c}{OpenAI Whisper} & \multicolumn{2}{c}{OWSM (ours)} \\
    \cmidrule(lr){2-3}
    \cmidrule(lr){4-5}
    & small & medium & v2 & v3\\
    \midrule
    1 & 4.4 & \textbf{3.8} & 7.2 & 9.2 \\
    5 & 4.2 & \textbf{3.8} & 6.6 & 7.6 \\
    \bottomrule
  \end{tabular}
  }
  \vskip -0.2in
\end{table}
\endgroup

\begingroup
\begin{table*}[t]
  \caption{Examples of ASR on 30-second audio segments, generated by OWSM v2 using greedy search. Utterances can be segmented in different ways, but the predicted timestamps are usually accurate. Differences between the reference and prediction are marked in \textcolor{red}{red}.}
  \label{tab:longform-examples}
  \centering
  \resizebox {\linewidth} {!} {
  \begin{tabular}{p{0.01\linewidth}p{0.5\linewidth}p{0.5\linewidth}}
    \toprule
    \# & Groundtruth from the dev set of MuST-C v2 & Prediction by OWSM v2 \\
    \midrule
    1 & $<$en$>$$<$asr$>$$<$0.00$>$ I'm going to talk today about energy and climate.\textcolor{red}{$<$3.50$>$$<$4.28$>$} And that might seem a bit surprising, because my full-time work at the foundation is mostly about vaccines and seeds, about the things that we need to invent and deliver to help the poorest two billion live better lives.\textcolor{red}{$<$18.38$>$$<$19.64$>$} But energy and climate are extremely important to these people\textcolor{red}{;} in fact\textcolor{red}{,} more important than to anyone else on the planet.$<$28.52$>$ & $<$en$>$$<$asr$>$$<$0.00$>$ I'm going to talk today about energy and climate.\textcolor{red}{$<$3.52$>$$<$4.26$>$} And that might seem a bit surprising, because my full-time work at the foundation is mostly about vaccines and seeds, about the things that we need to invent and deliver to help the poorest two billion live better lives.\textcolor{red}{$<$18.40$>$$<$19.62$>$} But energy and climate are extremely important to these people\textcolor{red}{,} in fact more important than to anyone else on the planet.$<$28.52$>$ \\
    \midrule
    2 & $<$en$>$$<$asr$>$$<$0.00$>$ Several years ago here at TED, Peter Skillman introduced a design challenge called the \textcolor{red}{marshmallow challenge.$<$5.60$>$$<$5.80$>$ And} the \textcolor{red}{idea's} pretty simple\textcolor{red}{:} Teams of four have to build the tallest \textcolor{red}{free-standing} structure out of \textcolor{red}{20 sticks} of spaghetti, one yard of tape, one yard of string and a marshmallow.\textcolor{red}{$<$16.52$>$$<$16.52$>$} The marshmallow has to be on top.\textcolor{red}{$<$18.18$>$}$<$18.54$>$ And\textcolor{red}{,} though it seems really simple, it's actually pretty hard because it forces people to collaborate very quickly.$<$25.04$>$\textcolor{red}{$<$25.42$>$} And so\textcolor{red}{,} I thought this was an interesting idea, and I incorporated it into a design workshop.$<$29.72$>$ & $<$en$>$$<$asr$>$$<$0.00$>$ Several years ago here at TED, Peter Skillman introduced a design challenge called the \textcolor{red}{Marshmellow Challenge, and} the \textcolor{red}{idea is} pretty simple\textcolor{red}{.$<$7.32$>$$<$7.50$>$} Teams of four have to build the tallest \textcolor{red}{freestanding} structure out of \textcolor{red}{26} of spaghetti, one yard of tape, one yard of string and a marshmallow.\textcolor{red}{$<$16.50$>$$<$16.54$>$} The marshmallow has to be on top.\textcolor{red}{$<$18.20$>$}$<$18.54$>$ And though it seems really simple, it's actually pretty hard because it forces people to collaborate very quickly.$<$25.04$>$\textcolor{red}{$<$25.44$>$} And so I thought this was an interesting idea, and I incorporated it into a design workshop.\textcolor{red}{$<$30.00$>$} \\
    \bottomrule
  \end{tabular}
  }
\end{table*}
\endgroup

\begingroup
\setlength{\tabcolsep}{2pt}
\begin{table}[t]
  \caption{BLEU \% ($\uparrow$) of speech translation. OpenAI Whisper supports any-to-English translation. OWSM can support more directions. The sizes of training sets (in hours) are also provided.}
  \label{tab:translation}
  \centering
  \resizebox {\linewidth} {!} {
  \begin{tabular}{ccccccccccccc}
    \toprule
    \multirow{2}{*}{Dataset} & \multirow{2}{*}{Source} & \multirow{2}{*}{Target} & \multicolumn{3}{c}{OpenAI Whisper} & \multicolumn{2}{c}{OWSM v2} & \multicolumn{2}{c}{OWSM v3} \\
    \cmidrule(lr){4-6}
    \cmidrule(lr){7-8}
    \cmidrule(lr){9-10}
    & & & hours & small & medium & hours & result & hours & result \\
    \midrule
    \multirow{5}{*}{MuST-C} & \multirow{5}{*}{English} & German & \multicolumn{3}{c}{\multirow{5}{*}{NA}} & 14k & \textbf{28.5} & 14k & 27.9\\
    & & Chinese & &&& 14k & 20.5 & 14k & \textbf{20.7} \\
    & & Japanese & &&& 1.0k & \textbf{10.5} & 1.0k & 9.4 \\
    & & Spanish & &&& 0.5k & \textbf{23.4} & 0.5k & 22.5\\
    & & French & &&& 0.5k & \textbf{28.5} & 0.5k & 26.2\\
    \midrule
    \multirow{5}{*}{CoVoST} & German & \multirow{5}{*}{English} & 4.3k & 26.2 & \textbf{34.8} & 0.2k & 18.6 & 0.2k & 18.0 \\
    & Chinese & & 12k & 6.3 & \textbf{13.6} & 0.01k & 3.0 & 0.01k & 3.3 \\
    & Japanese & & 8.9k & 15.9 & \textbf{22.9} & 0.001k & 0.1 & 0.001k & 0.1 \\
    & Spanish & & 6.7k & 34.2 & \textbf{40.2} & 0.1k & 24.9 & 0.1k & 22.7 \\
    & French & & 4.5k & 27.8 & \textbf{34.8} & 0.3k & 26.0 & 0.3k & 23.7 \\
    \bottomrule
  \end{tabular}
  }
\end{table}
\endgroup

\begingroup
\begin{table}[t]
  \caption{Accuracy \% ($\uparrow$) of language identification. OWSM~v3 supports 151 languages, whereas Whisper supports 99 languages.}
  \label{tab:lid}
  \centering
  \resizebox {0.8\linewidth} {!} {
  \begin{tabular}{cccc}
    \toprule
    \multirow{2}{*}{Dataset} & \multicolumn{2}{c}{OpenAI Whisper} & \multicolumn{1}{c}{OWSM (ours)} \\
    \cmidrule(lr){2-3}
    \cmidrule(lr){4-4}
    & small & medium & v3\\
    \midrule
    FLEURS & 53.1 & 54.8 & \textbf{81.4} \\
    \bottomrule
  \end{tabular}
  }
\end{table}
\endgroup

\begingroup
\setlength{\tabcolsep}{3pt}
\begin{table}[t]
  \caption{WER/CER \% ($\downarrow$) of OWSM v3 using different decoding algorithms in ESPnet.}
  \label{tab:decoding}
  \centering
  \resizebox {\linewidth} {!} {
  \begin{tabular}{ccccc}
    \toprule
    Dataset & Metric & CTC & Attention & Joint CTC/attention \\
    \midrule
    Common Voice en & \multirow{8}{*}{WER} & 18.6 & 14.5 & \textbf{12.9} \\
    FLEURS en & & 17.3 & 10.9 & \textbf{9.7} \\
    LibriSpeech test-clean & & 4.5 & 2.7 & \textbf{2.6} \\
    LibriSpeech test-other & & 8.1 & 6.0 & \textbf{5.4} \\
    Switchboard eval2000 & & 19.4 & 17.2 & \textbf{16.6} \\
    TEDLIUM test & & 6.7 & 4.8 & \textbf{4.7} \\
    VoxPopuli en  & & 12.3 & 9.2 & \textbf{8.7} \\
    WSJ eval92 & & 32.0 & 13.4 & \textbf{11.4} \\
    \midrule
    AISHELL-1 test & \multirow{2}{*}{CER} & 9.2 & 7.1 & \textbf{6.5} \\
    ReazonSpeech test & & 17.0 & 11.3 & \textbf{10.3} \\
    \bottomrule
  \end{tabular}
  }
\end{table}
\endgroup

\section{Experiments}

\subsection{English speech recognition}
\label{subsec:exp-en-asr}

Table~\ref{tab:en-asr} presents word error rates (WER) on standard English ASR benchmarks. Greedy search is employed without any external language models. To ensure fair comparison, we prepare all test data in ESPnet and evaluate Whisper in the same setup instead of reporting results from their paper~\cite{openai-whisper}. The text is normalized using the English or basic normalizer provided by Whisper. Whisper large is not included since it is significantly larger than the other models.
Although many public ASR corpora are combined, our English training data is still significantly smaller than that of Whisper (73k vs 438k hours). However, our OWSM models achieve competitive results in most benchmarks. OWSM models even outperform Whisper on LibriSpeech and Switchboard.

By comparing different versions of OWSM, we observe that its English ASR capability is largely improved from v1 to v2, demonstrating the effectiveness of scaling up in terms of the number of model parameters and the amount of training data. However, OWSM~v3 does not show a consistent improvement over v2 in all benchmarks. OWSM~v3 achieves lower WERs on Switchboard and VoxPopuli test sets, likely because their training sets are newly added (see Section~\ref{subsec:dataprep}). OWSM~v3 has slight degradations on LibriSpeech and a large degradation on WSJ. This is probably due to the shift of data distributions from v2 to v3. As shown in Table~\ref{tab:data-model-stats}, our v3 dataset contains significantly more languages compared to v2 (151 vs 23), but the model size is only slightly increased (889M vs 712M). Hence, the model has to adjust its capacity from English to other languages or from one type of speech to another type. This issue might be mitigated with larger models and more diverse data. We will explore it in the future. Please refer to the last paragraph in Section~\ref{subsec:trainingtips} for more discussions.

We have also investigated the inference speed. Specifically, we select 50 utterances of 30 seconds from our prepared TEDLIUM dev set, and decode OWSM v3 with greedy search using a single NVIDIA A40 GPU. The average decoding time for each 30-second utterance is $2.3$ seconds.

\subsection{Multilingual speech recognition}
\label{subsec:multilingual-asr}

Table~\ref{tab:multilingual-asr} shows the ASR results on multilingual benchmarks. In general, OpenAI Whisper achieves better performance than our OWSM, because Whisper employs significantly more training data in all languages except Japanese. For Japanese, OWSM v3 outperforms Whisper by a large margin (CER: 11.3 vs 25.3) thanks to the larger amount of training data (19k vs 7k hours) from ReazonSpeech~\cite{reazonspeech}. Notably, OWSM v2 achieves the best results on the English and Chinese test sets from Multilingual LibriSpeech and AISHELL, respectively, despite being trained on less data.

The trend across different versions of OWSM is consistent with that in Section~\ref{subsec:exp-en-asr}. OWSM v2 is drastically improved compared to v1 in all languages, which verifies the benefits of scaling up. OWSM v3 outperforms v2 in a few languages but achieves comparable or slightly worse results in the others. Again, this is likely because the model needs to adjust its capacity to support much more languages in v3.

\subsection{Long-form speech recognition}

Similar to Whisper, OWSM performs long-form ASR by consecutively transcribing 30-second audio segments and shifting the window based on predicted timestamps. Table~\ref{tab:longform-asr} presents the long-form ASR results on the TEDLIUM test set, where each input audio is an unsegmented long talk. OWSM~v2 achieves 7.2\% WER with greedy decoding and 6.6\% WER with beam search. Whisper models achieve lower WERs in both cases, likely because: (1) their training set is larger; (2) their data, collected from the Internet, is originally in a long form, which can be more realistic than ours; (3) they apply various heuristics to improve the timestamp prediction and also the quality of text (see Section 4.5 in their official report~\cite{openai-whisper}). In our future work, we will explore more strategies to enhance the long-form performance.

Table~\ref{tab:longform-examples} shows two examples from TED talks, where timestamps are generated along with text tokens. Although utterances can be segmented in different ways, the boundaries predicted by OWSM are usually very close to the reference.

\subsection{Speech translation}
\label{subsec:exp-st}

Table~\ref{tab:translation} compares different models on two ST benchmarks: MuST-C (English-to-X) and CoVoST (X-to-English). Whisper only supports the latter, while OWSM supports both directions.

OWSM models achieve notable results on MuST-C thanks to the sufficient amount of training data (more than 500 hours for each language). The BLEU scores of Chinese and Japanese are lower than those of European languages even with enough training data. This indicates that the model has difficulty in translating between very different languages. 

On CoVoST, the performance of OWSM ranges across language pairs as the amount of training data varies from 1 to 300 hours. On Chinese and Japanese, OWSM outputs have low intelligibility while on the European languages OWSM outputs are moderately intelligible.
On the other hand, Whisper is trained on 4k to 12k hours and thus achieves greater BLEU scores on X-to-English in general.

Similar to the findings in Section~\ref{subsec:exp-en-asr} and Section~\ref{subsec:multilingual-asr}, OWSM~v3 shows comparable or slightly worse performance than OWSM~v2. This is because OWSM~v3 employs almost the same amount of ST data but it has to recognize drastically more languages (see Table~\ref{tab:data-model-stats}). Some of its capacity needs to be assigned to these additional languages.

\subsection{Language identification}

As described in Section~\ref{subsec:multitask-data-format} and Figure~\ref{fig:overview}, OWSM predicts a language token at the beginning of decoding, which effectively performs the LID task. Table~\ref{tab:lid} compares Whisper and OWSM on the FLERUS test set prepared in ESPnet. OWSM v3 achieves a top-1 accuracy of 81.4\%, which outperforms Whisper small and medium by a large margin. This is because OWSM v3 utilizes the training data from Common Voice and FLEURS, containing 151 languages in total, whereas Whisper supports 99 languages that only cover a subset of the languages in FLEURS. Nevertheless, this result demonstrates that OWSM has a strong capability in speech classification although it is designed as a sequence-to-sequence model.

\subsection{Comparison of decoding algorithms}

OWSM is compatible with various decoding algorithms in ESPnet. Table~\ref{tab:decoding} compares three commonly used algorithms: CTC only (greedy), attention only (greedy), and joint CTC/attention (with beam size 10 and CTC weight 0.3).
Beam search with joint CTC/attention achieves the best results in all test sets. Attention-only decoding outperforms CTC-only, indicating that the decoder has strong capacity.

\section{Discussions and future directions}

This work serves as an exploratory endeavor in reproducing Whisper-style training using open-source resources. Moving forward, we will delve into the following directions.

Firstly, the current OWSM still falls behind Whisper in many benchmarks, likely because: (1) OWSM supports more languages and more translation directions, which increases the difficulty of multitask learning; (2) our training set is significantly smaller than that of Whisper in nearly all languages and tasks; (3) we directly leverage public ASR and ST corpora which may be less diverse than Whisper's data collected from the Internet. These issues can probably be addressed by utilizing more advanced encoder~\cite{conformer, branchformer, ebranchformer, is23-ebranchformer} or decoder~\cite{state-space-decoder} architectures, collecting more diverse ASR and ST data from public sources, and incorporating self-supervised speech representations~\cite{wav2vec2, hubert} as in Google USM~\cite{google-usm}.

Secondly, we plan to incorporate additional speech processing tasks into the multitask framework, including spoken language understanding and speech generation based on discrete representations, thereby working towards the development of ``universal speech models''.

Thirdly, these large pre-trained models are unsuitable for deployment in real-world applications. Various compression techniques~\cite{distilhubert, parp, ssl-pruning, is23-dphubert, han2021dynamicsurvey, i3d-dynamic-models} can be applied to reduce the model size and computation.

Fourthly, OWSM provides a valuable testbed for investigating and exploring various machine learning problems such as data imbalance, continual learning~\cite{continual-learning}, adversarial robustness~\cite{robustness}, and machine unlearning~\cite{machine-unlearning}.

\section{Conclusion}

This work presents OWSM, which reproduces Whisper-style training using an open-source toolkit and publicly available data. OWSM follows the multitask framework of OpenAI Whisper, but extends it to support more translation directions. Several strategies are developed to improve efficiency. We will open-source all scripts for data preparation, training, inference, and scoring as well as pre-trained models and training logs. We believe this can promote transparency and facilitate advancements in the large-scale pre-training of speech models.

\section{Acknowledgements}
We use PSC Bridges2 and NCSA Delta via ACCESS allocation CIS210014, supported by National Science Foundation grants \#2138259, \#2138286, \#2138307, \#2137603, and \#2138296.

\bibliographystyle{IEEEbib}
\bibliography{refs}

\end{document}